\documentclass[pmlr, ruled, linesnumbered, boxed]{jmlr}

\theorembodyfont{\upshape}
\theoremheaderfont{\scshape}
\theorempostheader{:}
\theoremsep{\newline}

\jmlryear{2022}
\jmlrworkshop{LearnAut}

\usepackage{xcolor}
\usepackage{colortbl}

\newcommand{\saat}{SAAT }
\newcommand\mybar{\kern1pt\rule[-\dp\strutbox]{.7pt}{\baselineskip}\kern1pt}

\title[Robust Attack Graph Generation]{Robust Attack Graph Generation}

\author{\Name{Dennis Mouwen} \Email{D.Mouwen@student.tudelft.nl}\\
\Name{Sicco Verwer} \Email{S.E.Verwer@tudelft.nl}\\
\Name{Azqa Nadeem} \Email{azqa.nadeem@tudelft.nl}\\
\addr Delft University of Technology
}

\begin{document}

\maketitle

\begin{abstract}
We present a method to learn automaton models that are more robust to input modifications. It iteratively aligns sequences to a learned model, modifies the sequences to their aligned versions, and re-learns the model. Automaton learning algorithms are typically very good at modeling the frequent behavior of a software system. Our solution can be used to also learn the behavior present in infrequent sequences, as these will be aligned to the frequent ones represented by the model. We apply our method to the SAGE tool for modeling attacker behavior from intrusion alerts. In experiments, we demonstrate that our algorithm learns models that can handle noise such as added and removed symbols from sequences. Furthermore, it learns more concise models that fit better to the training data.
\end{abstract}


\section{Introduction}
\label{sec:introduction}
Most organizations and companies use an Intrusion Detection System (IDS) as the first line of defense against cyber attacks. The biggest limitation of these systems is the huge amount of alerts they generate. Alert Correlation techniques are used to reduce and group alerts belonging to the same attack stage \citep{Alserhani2015AlertCA, SALAH20131289, Sadoddin}. They can show \textit{what} an attacker did, but not \textit{how} they exploited the infrastructure. 
Attack Graphs (AG) are models that visually describe attacker strategies. While a lot of research has been done on AG generation, most methods that generate AGs rely heavily on expert knowledge~\citep{10.1145/586110.586144,Alserhani2015AlertCA,ning2004building},
which is both expensive and time-consuming, or published vulnerability reports~\citep{ou2005mulval,hu2020attack,roschke2011new,gao2018exploring},
which are always one step behind attackers. Alert-based AG generation methods only use (IDS) alerts. 
To the best of our knowledge, SAGE~\citep{azqa} is the first and only method for generating Alert-based AGs without prior knowledge. 
SAGE uses a suffix-based probabilistic deterministic finite automaton (S-PDFA) to learn and model these attacks. This model accentuates infrequent severe alerts, without discarding low-severity alerts. After the S-PDFA is learned, it is used to recognize known attacks in new incoming alerts. 

A pitfall of SAGE however, is that these alerts must exactly follow a path in the state machine. In real-world scenarios, this is usually not the case, e.g., an alert is missing due to a new attacker technique or a defective IDS. Next to that, infrequent alerts can still be missed due to the nature of the statistical tests used in automaton learning algorithms. These are problems for probabilistic automaton learning algorithms and lead to missing attack paths in the model. 
SAGE uses the FlexFringe \citep{flexfringe} implementation of Alergia~\citep{carrasco1994learning} to learn probabilistic automata. This implementation has some mechanisms to deal with infrequent traces such as the use of sink states. These leave the infrequent parts of the prefix tree untouched because the statistical tests performed on them are unreliable, making the learning algorithm less likely to draw incorrect inferences. As a consequence, it learns nothing from infrequent traces. 

In this paper, we propose a different strategy for dealing with them, which is to align these sequences to the frequent parts of the learned automaton. We call our method Sequence-Automaton Alignment Tool (SAAT). SAAT computes an \textit{alignment} between a sequence and an automaton, and computes a (normalised) score for each alignment produced. The average score (over all training data) can be used to determine how accurate the model captures the data. 
An alignment can introduce new symbols or remove old ones, thus dealing with missing or misplaced alerts. After alignment, we build a re-aligned data set and rerun the main learning algorithm. By iteratively performing this process, we obtain smaller and better aligned models than the original ones produced by FlexFringe. We demonstrate \saat's effectiveness on the CPTC-2018 security testing competition dataset\citep{cptc-2018}, which results in a 49\% smaller model, while actually increasing how well the model describes the data. 
Our main contributions are:
\begin{enumerate}
    \item We propose Sequence-Automaton Alignment Tool (SAAT), which computes an \textit{alignment} between a sequence and an automaton. Iteratively using the alignments results in a smaller and more interpretable model.
    \item We provide a definition of the normalised alignment score, a metric that quantifies how well a model fits the data.
    \item We show in experiments that SAAT reduces the model size by almost a half, while improving the normalised score.
\end{enumerate}


\section{Related Work}
\label{sec:related-work}

Learning automata (state machines) from traces is a grammatical inference~\citep{higuera2010book} problem where traces are modeled as the words of a language, and the goal is to find a model for this language, e.g., a (probabilistic) deterministic finite state automaton. Learning such models from software trace data~\citep{cook1998discovering} is not new and has been used for analyzing different types of complex software systems such as web-services~\citep{bertolino2009automatic,ingham2007learning}, X11 programs~\citep{ammons2002mining}, communication protocols~\citep{comparetti2009prospex,antunes2011reverse,fiterau2020analysis}, Java programs~\citep{cho2011mace}, and malicious software~\citep{cui2007discoverer}. Popular tools for leaning form software trace data are CSight~\citep{beschastnikh2014inferring} and MINT~\citep{walkinshaw2016inferring}. SAGE~\citep{azqa} is the first tool to use automaton learning to infer attack graphs from intrusion data.



Although alignment is not common in automaton learning, in process mining~\citep{van2012process} it is frequently used in process conformance testing~\citep{rozinat2010process,carmona2018conformance}, which uses sequence alignment to determine the fit of a process model to a log of trace data. We perform a similar process but using a new sequence-to-model algorithm based on~\citep{sofia}. We will refer this algorithm as the TS algorithm. It is a dynamic programming algorithm based on the Needleman-Wunsch sequence alignment algorithm \citep{NEEDLEMAN1970443}.

\section{Sequence-Automaton Alignment Tool}
\label{sec:saat}
Sequence-Automaton Alignment Tool (SAAT) is developed around the concept of an alignment between a sequence and an automaton. Recall that a sequence is a list of symbols, e.g., or [apple, banana, cherry], and an automaton (or finite state machine) is a directed graph with labels on the edges. In this work, an alignment is a list of \textit{matched}, \textit{skipped}, or \textit{added} edges. A matched edge describes both a symbol in the sequence, and an edge in the model, while added and skipped edges describe the differences between the sequence and the model.
\saat performs \textit{global} alignment, meaning that the full sequence will be aligned on the automaton.

\begin{figure}[t]
\floatconts
  {fig:example_alignment_matrix}
  {\caption{An example automaton and a matched sequence. The number in parentheses is the level of the node.}}
  {\subfigure[Example automaton $A$. ]{\label{fig:alignment-matrix-automaton}%
      \includegraphics[width=0.28\linewidth]{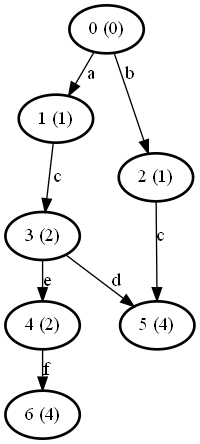}}%
    \qquad \qquad
    \qquad \qquad
    \subfigure[Matched sequence $a \rightarrow b \rightarrow$ $e \rightarrow d \rightarrow f$]{\label{fig:subtree-matched-trace}%
      \includegraphics[width=0.28\linewidth]{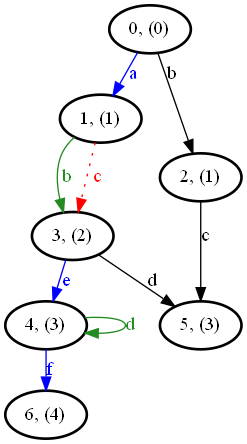}}
  }
\end{figure}

\begin{table}[t]
\begin{center}
  {%
    \begin{tabular}{|l|l|l|l|l|l|l|l|l|l|}
\hline
Index & Level & Transition        & Symbol &                           & 0: a                      & 1: b                      & 2: e                      & 3: d                      & 4: f                      \\ \hline
0     &       &                   &        & 0 & -2                        & -4                        & -6                        & -8                        & -10                       \\ \hline
1     & 0     & $0 \rightarrow 1$ & a      & -2                        & \cellcolor[HTML]{9AFF99}4 & \cellcolor[HTML]{DAE8FC}2 & 0                         & -2                        & -4                        \\ \hline
2     & 1     & $1 \rightarrow 3$ & c      & -4                        & 2                         & 0                         & -2                        & -4                        & -6                        \\ \hline
3     & 2     & $3 \rightarrow 4$ & e      & -6                        & 0                         & -2                        & \cellcolor[HTML]{9AFF99}4 & \cellcolor[HTML]{DAE8FC} 2                         & 0                         \\ \hline
4     & 2     & $3 \rightarrow 5$ & d      & -6                        & 0                         & -2                        & -4                        & 2 & 0                         \\ \hline
5     & 3     & $4 \rightarrow 6$ & f      & -8                        & -2                        & -4                        & 2                         & 0                         & \cellcolor[HTML]{9AFF99}6 \\ \hline
\end{tabular}
  }
  {\caption{\label{tab:example-alignment-matrix}The alignment matrix corresponding to the matched sequence of \figureref{fig:subtree-matched-trace} with Linear Scoring \hbox{(\textit{MATCH}=4, \textit{MISMATCH}=-4, \textit{GAP}=-2)}. Green cells show a \textit{MATCH}, blue cells show a \textit{GAP}. The highlighted cells are the maximum value in each column. }}%
\end{center}
\end{table}

\subsection{Computing the Alignment Matrix}
The alignment matrix is computed in similar manner as the TS algorithm~\citep{sofia}. It is computed for every sequence, and can afterwards be used to find to which edge a symbol of the sequence is best matched. The matrix has size $(m + 1) \times (n+1)$, with $m = $ the amount of edges of the automaton, and $n = $ the length of the sequence. The rows thus represent edges, while the columns represent the symbols. \tableref{tab:example-alignment-matrix} shows the alignment matrix for the automaton in \figureref{fig:alignment-matrix-automaton} and the sequence \hbox{$a \rightarrow b \rightarrow e \rightarrow d \rightarrow f$}.
The initial values (top row and left-most column) are the rows multiplied by the level (depth) of an edge, and the columns by the index of the symbol. The content of the cells are computed using scores for a \textit{MATCH} and penalties for a \textit{MISMATCH} and \textit{GAP}. Each cell $c_{i,j}$ corresponds to triggering transition $i$ (row) with the label $l_j$ at index $j$ (column). Each transition is a tuple $(s_i, t_i, l_i)$ of a source state $s_i$, a target state $t_i$, and a transition label $l_i$. A cell $c_{i,j}$ is updated with the maximum value of:
\begin{eqnarray*}
c_{k,j-1} + \textit{MATCH} \text{ if } s_i = t_k \text{ and } l_j = l_k \\
c_{k,j-1} - \textit{MISMATCH} \text{ if } s_i = t_k \text{ and } l_j \not= l_k \\
c_{k,j-1} - \textit{GAP} \text{ if } t_i = t_k\\
c_{k,j} - \textit{GAP} \text{ if } s_i = t_k
\end{eqnarray*}
Thus, we follow a transition if one with the correct label exists and add a score. We can also follow a transition with the wrong label but then add a large penalty. A smaller penalty is given to staying in the current state (skipping a symbol from the sequence) or following a transition without advancing the sequence index (skipping a transition from the model). For more details on the alignment algorithm, we refer to~\citep{sofia}. We note that the use of $c_{k,j}$ in the update gives problems in the case of loops and for that reason the algorithm operates per level, essentially unrolling loops.

\subsection{Score of the Alignment}
\label{sec:normalised-score}
The alignment system gives a score to the computed alignment. This score gives an indication of how well the sequence is aligned to the automaton. However, using the score as-is always gives longer sequences a higher score, as more points can be scored. Thus, a normalised score $S_{norm}$ is used: 
%
\begin{align*}
    S_{max}  &= n * MATCH\\
    S_{min}  &= n * MISMATCH\\
    S_{norm} &= \frac{s - S_{min}}{S_{max} - S_{min}}
\end{align*}
where $s$ the score of the alignment and $n$ the length of the sequence. Since \textit{MATCH} is the maximum score possible for a single symbol, it is clear that a score of 1 can be reached by having all symbols in the sequence aligned to a transition in the automaton. 
The normalised score can be used as a metric to quantify how well a sequence fits the model. We use it to quantify how well a model fits the data.
\section{Iterative Models}
\label{sec:iterative-models}
\saat computes the differences (or alignments) between the training data and the model. From these differences, it can be deduced where the model falls short and how it can be improved. The differences are used to generate a new Flexfringe training file, resulting in a new and  possibly improved model. By iteratively doing this process, the model converges to an optimum.
The new training data is generated according to a set of heuristics.
The input is a list of alert sequences. The general outline of this method is as follows:

\begin{enumerate}
\item Aggregate the alerts sequences into episode sub-sequences (ESS), and generate a Flexfringe training file from these ESS.
\item Use Flexfringe to learn a model from this training file.
\item Compute the alignment between the learned model and the \textbf{original} ESS.
\item Generate a new Flexfringe training file from these matched sequences.
\item Repeat steps 2 to 4 with the new training file, until convergence. 
\end{enumerate}

\begin{figure}[htbp]
    \makebox[\textwidth][c]{\includegraphics[width=1.0\textwidth]{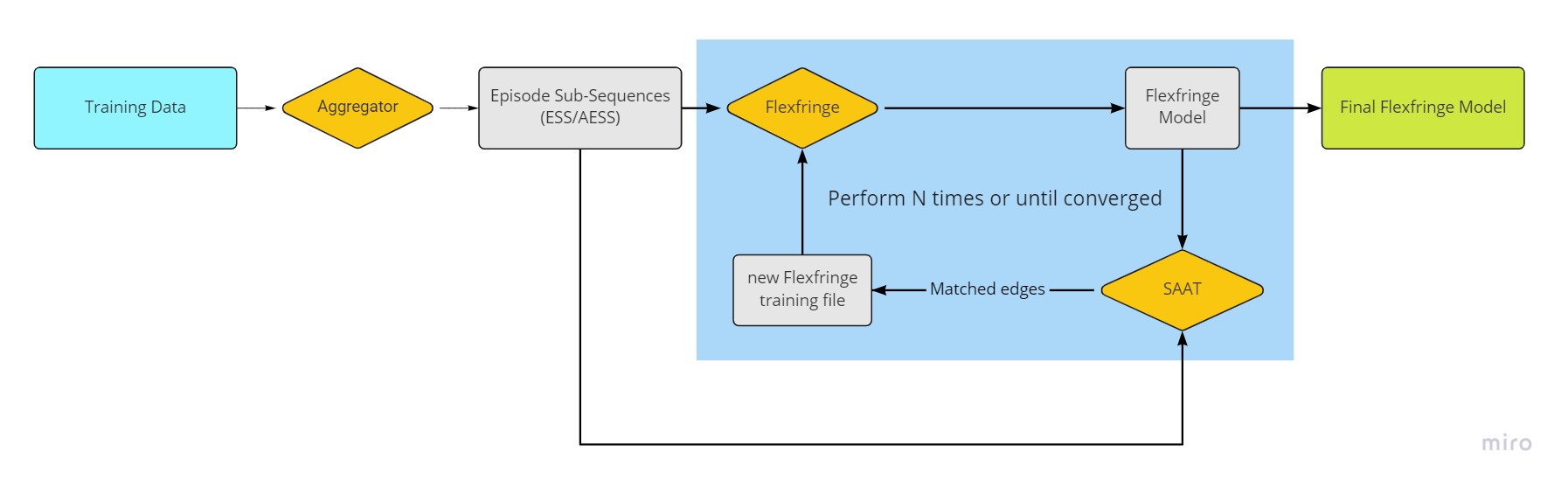}}%
    \caption{Overview of iterative models}
    \label{fig:iterative-overview}
\end{figure}
Initially, only matched edges will be used for the new training data. For example, if there is only one ESS, with $a \rightarrow b \rightarrow c \rightarrow d$, of which the edges $a$, $c$, and $d$ are matched (and $b$ is skipped), the new training data will only contain $a \rightarrow c \rightarrow d$.
The maximum number of iterations will be set to 16. This number should be large enough to converge, while keeping the training time low enough.

\subsection{Model Evaluation}
To determine the quality of the learned iterative models, the normalised score (\sectionref{sec:normalised-score}) is computed for each of the matched sequences. If there is an error, the sequence will get the score 0. The average normalised score is computed over all matched sequences, and indicates how well the model fits the data. We opt for this method instead of other metrics like perplexity, since the sequences can be only partially present in the model. The probability of such a sequence would be 0, but the normalised score will give a score between 0 and 1. 
The average normalised score is saved for each iteration. When the average normalised score is the same for at least 3 iterations, we can conclude that the score has converged. From all iterations, the model with the best average normalised score is picked as the final model, and is used to generate the attack graphs.

Next to the normalised score, the size of the model is taken into account. A smaller model can be less descriptive, but also easier to interpret. As such, we favour smaller models over bigger models. 
The original model (from SAGE), trained with standard parameters, has an average normalised score of 0.656, and consists of 256 nodes.

\subsection{Generating New Training Data}
\label{sec:heuristics}
The new training data is generated from the alignments of the previous iteration. For each alignment, only the \textit{matched} edges are used as training data for the next iteration. Next to that, two heuristics are used to improve the training data:

\paragraph{Remove Sequential Duplicates}
To make the model more concise, sequential duplicates in the training data are removed. As \saat can recover these symbols in the form of added edges, this change does not affect the resulting attack graphs.

\paragraph{Add missing Sequences}
If the objective of a sequence (the first symbol) is not present in the model, the sequence can never be matched. When this happens, this sequence is added back to the training data.

\section{Dataset \& Experimental Setup}
\label{sec:evaluation}
To verify and validate the capabilities of \saat, perform two experiments:
\begin{enumerate}
    \item Algorithm verification: We show that \saat works as intended by applying simple modifications to a set of sequences.
    \item Learning comparison: We compare the quality of models learned by iterative \saat and SAGE, with regards to normalised score, model size, and interpretability. 
\end{enumerate}

\subsection{CPTC-2018 Dataset}
Security penetration testing competitions provide an ideal environment for multi-stage attacks.  We apply \saat to the CPTC-2018 dataset\citep{cptc-2018}. This public IDS alert dataset contains Suricata alerts generated by six different teams whose objective is to compromise a network infrastructure. Their methods may differ, but their goals are the same. This makes the dataset interesting for automata learning, as their strategies are not equal but do overlap. 
In \citep{munaiah} a more detailed explanation on this infrastructure and the dataset is given. 
Before the alerts are used, they are parsed and aggregated into Episode Sequences by SAGE. The automata are generated using Flexfringe, we refer to~\citep{verwer2022flexfringe} for an overview of its inner workings. As IDS are used as data, which is positive data, the Alergia state merging algorithm \citep{alergia} is used. The parameters used for Flexfringe are the same as in \citep{azqa}.


\subsection{Algorithm Verification Experiments}
The focus lays on whether \saat can recover the modifications that were made to the data. They will be evaluated on a learned model of the CPTC-2018 dataset. A test case will be considered as successful when \saat is able to correctly recover the original trace. For example, if $c$ is removed from the trace $[a \rightarrow b \rightarrow c \rightarrow d]$, \saat should be able to recover $[a \rightarrow b \rightarrow d]$ back to $[a \rightarrow b \rightarrow c \rightarrow d]$, including the correct node ids. If any of the symbols is wrong, the test is not correct. We test 7 test cases: 
\begin{itemize}
    \item REMOVE-1 and REMOVE-2, where 1 or 2 symbols are removed from the sequence. 
    \item ADD-1 and ADD-2, where 1 or 2 random symbols are added to the sequence. 
    \item MODIFY-1 and MODIFY-2, where 1 or 2 symbols are modified. 
    \item SWAP-1, where 2 sequential symbols are swapped.
\end{itemize}
These test cases have been chosen to mimic real-world scenarios in which an attack (partially) changes. For example, an IDS might miss an alert due to a new attack technique which cannot be detected yet. Or an attacker might swap two steps if they are interchangeable. 
To generate the modified sequences, all possible attack paths in the model are retrieved. For each attack path, all possible permutations possible for the current test case are evaluated. For example, for REMOVE-1, if the attack path is $n$ nodes long, then there are $n-2$ possible permutations: One for every possible removed node.

\subsection{Learning Comparison}
Two models are learned using iterative \saat and SAGE. Every training data sequence will be aligned to the models. For each aligned trace, the normalised score is computed. The final score is the average of all normalised scores. We have chosen for this metric over perplexity because a) perplexity is hard to compare over different datasets and implementations, and b) the normalised score better expresses partially correct solutions.

\subsection{Results}
In \tableref{tab:cptc-2018-results}, the results for the algorithm verification experiments are shown. \saat is able to recover from these simple modifications, and works as intended. MODIFY-2 has 6 incorrect cases. Most of these are cases where two sequential symbols are modified. The next symbol after these modified symbols is incorrectly labeled as \textit{added} instead of \textit{matched}. 
In \tableref{tab:iterative-results}, the results of comparing the learned models of iterative \saat and SAGE can be seen. The model size has been greatly reduced  from 256 nodes to 131.3 nodes, a 49\% decrease. This makes sense, as the infrequent traces are made to look like frequent ones, reducing the size of the prefix tree. Furthermore, the average normalised score has actually improved from 0.656 to 0.706. Note that this is computed with respect to the original training data. The \saat model is therefore better at expressing the learned data with much fewer states. 

\begin{table}[htbp]
    \centering
    \begin{tabular}{|l|r|r|c|}
        \hline
        test case & number of tests & correct & {accuracy}             \\ \hline
        REMOVE-1  & 464  & 464  & \cellcolor[HTML]{C0C0C0}\textbf{1.0} \\ \hline
        REMOVE-2  & 1855 & 1855 & \cellcolor[HTML]{C0C0C0}\textbf{1.0} \\ \hline
        ADD-1     & 464  & 464  & \cellcolor[HTML]{C0C0C0}\textbf{1.0} \\ \hline
        ADD-2     & 1855 & 1855 & \cellcolor[HTML]{C0C0C0}\textbf{1.0} \\ \hline
        MODIFY-1  & 464  & 462  & \cellcolor[HTML]{C0C0C0}\textbf{1.0} \\ \hline
        MODIFY-2  & 1855 & 1849 & \cellcolor[HTML]{C0C0C0}\textbf{0.996} \\ \hline
        SWAP-1    & 380  & 359  & \cellcolor[HTML]{C0C0C0}\textbf{0.944} \\ \hline
    \end{tabular}
    \caption{Results for the verification experiments of the CPTC-2018 dataset.}
    \label{tab:cptc-2018-results}
\end{table}

\begin{table}[hbtp]
  \begin{center}
  \begin{tabular}{|l|c|c|c|c|}
               & {Avg. Norm. Score} & {SD Norm. Score} & {Avg. Size} & {SD Size} \\ \hline
SAAT iterative & 0.706       & 0.014      & 131.3     & 8.825   \\
SAGE           & 0.656      & {-}        & 256       & {-} \\
\end{tabular}
\end{center}
  \caption{\label{tab:iterative-results}Result of average normalised score (over 10 experiments) and size of Iterative SAAT compared to SAGE, including standard deviation.}
\end{table}

\section{Conclusions and Future Work}
\label{sec:conclusions}
We present a new alignment process called \saat that can be used to learn smaller and more robust automaton models. Performing this operation in an iterated process can reduce the model size from 256 nodes to 131 nodes (a 49\% decrease), while actually improving how well the model fits the data. The normalised score can quantify the performance of a model and is used to evaluate the iterative process.

With this work, we present a new approach to deal with a common problem in state merging algorithms: what to do with infrequent states. FlexFringe labels these as sinks states, essentially ignoring them. By aligning these sequences to the frequent part of the learned model, we demonstrate how the data from these traces can be used during learning. Future work will focus on applying \saat to additional datasets, further utilizing the sink states for the alignment process, and improving the alignment algorithm to better handle cycles in the automaton.

\newpage

\bibliography{natbib.bib}






\end{document}